\documentclass{article}
\usepackage[nonatbib,preprint]{nips_2018}

\usepackage{cite}
\usepackage[T1]{fontenc}    % use 8-bit T1 fonts
\usepackage{hyperref}       % hyperlinks
\usepackage{url}            % simple URL typesetting
\usepackage{booktabs}       % professional-quality tables
\usepackage{amsfonts}       % blackboard math symbols
\usepackage{nicefrac}       % compact symbols for 1/2, etc.
\usepackage{microtype}      % microtypography
\usepackage{bm}
\usepackage{amsmath}
\usepackage{mathtools}

\usepackage{graphicx}
\graphicspath{ {images/} }
\newcommand{\bhline}[1]{\noalign{\hrule height #1}}
\usepackage{wrapfig}

\title{Rethinking the role of normalization and residual blocks for spiking neural networks}

\author{
  Shin-ichi Ikegawa\\
  Tokyo Research Center, Aisin\\
  \texttt{shinichi.ikegawa@aisin.co.jp} \\
  \And
  Ryuji Saiin\\
  AISIN SOFTWARE Co., Ltd.\\
  \texttt{ryuji.saiin@aisin-software.co.jp}\\
  \And
  Yoshihide Sawada\\
  Tokyo Research Center, Aisin\\
  \texttt{yoshihide.sawada@aisin.co.jp} \\
  \And
  Naotake Natori\\
  Tokyo Research Center, Aisin\\
  \texttt{naotake.natori@aisin.co.jp} \\
}

\begin{document}

\maketitle

\begin{abstract}
Biologically inspired spiking neural networks (SNNs) are widely used to realize ultralow-power energy consumption. However, deep SNNs are not easy to train due to the excessive firing of spiking neurons in the hidden layers. To tackle this problem, we propose a novel but simple normalization technique called postsynaptic potential normalization. This normalization removes the subtraction term from the standard normalization and uses the second raw moment instead of the variance as the division term. The spike firing can be controlled, enabling the training to proceed appropriating, by conducting this simple normalization to the postsynaptic potential. The experimental results show that SNNs with our normalization outperformed other models using other normalizations. Furthermore, through the pre-activation residual blocks, the proposed model can train with more than 100 layers without other special techniques dedicated to SNNs.
\end{abstract}

\section{Introduction}
\label{sec:1}
Recently, spiking neural networks (SNNs) \cite{Maass-1997} have attracted substantial attention due to ultra-low power consumption and high friendliness with hardware such as neuromorphic-chips \cite{Akopyan-2015, Davies-2018} and field-programmable gate array (FPGA) \cite{Maguire-2007}. In addition, SNNs are biologically more plausible than artificial neural networks (ANNs) because their neurons communicate with each other through spatio-temporal binary events (spike trains), similar to biological neural networks (BNNs). However, SNNs are difficult to train since spike trains are non-differentiable.

Several researchers have focused on the surrogate gradient to efficiently train SNNs \cite{Wu-2018, Zenke-2018, Shrestha-2018, Lee-2020, Zhang-2020}. The surrogate gradient is an approximation of the true gradient and is applied to the backpropagation (BP) algorithm \cite{Rumelhart-1986}. Recent studies have successfully trained deep SNNs using this method \cite{Fang-2021b}. However, it is still challenging to train deepened models due to the increasing difficulty of controlling spike firing.

To control the spike firing properly, we propose a novel and simple normalization: {\it postsynaptic potential normalization}. Contrary to the standard batch/layer normalizations, our normalization removes the subtraction term from the standard normalization and uses the second raw moment instead of the variance as the division term. We can automatically control the firing threshold of the membrane potential and spike firing by conducting this simple normalization to the postsynaptic potential (PSP). The experimental results on neuromorphic-MNIST (N-MNIST) \cite{Orchard-2015} and Fashion-MNIST (F-MNIST) \cite{Xiao-2017} show that SNNs with our normalization outperform other models using other normalizations. We also show that the proposed method can train the SNN consisting of more than 100 layers without other special techniques dedicated to SNNs.

The contributions of this study are summarized as follows.
\begin{itemize}
  \item We propose a novel and simple normalization technique based on the firing rate. The experimental results show that the proposed model can simultaneously achieve high classification accuracy and low firing rate.
  \item We trained deep SNNs based on the pre-activation residual blocks \cite{He-2016a}. Consequently, we successfully obtained a model with more than 100 layers without other special techniques dedicated to SNNs.
\end{itemize}

The remainder of the paper is organized as follows. In Sections \ref{sec:related-works} - \ref{sec:3}, we describe the related works, SNN used in this paper, and our normalization technique. Section \ref{sec:4} presents the experimental results. Finally, Section \ref{sec:5} presents the conclusion and future works.

\begin{figure}[t]
\begin{center}
\includegraphics[width=1.0\linewidth]{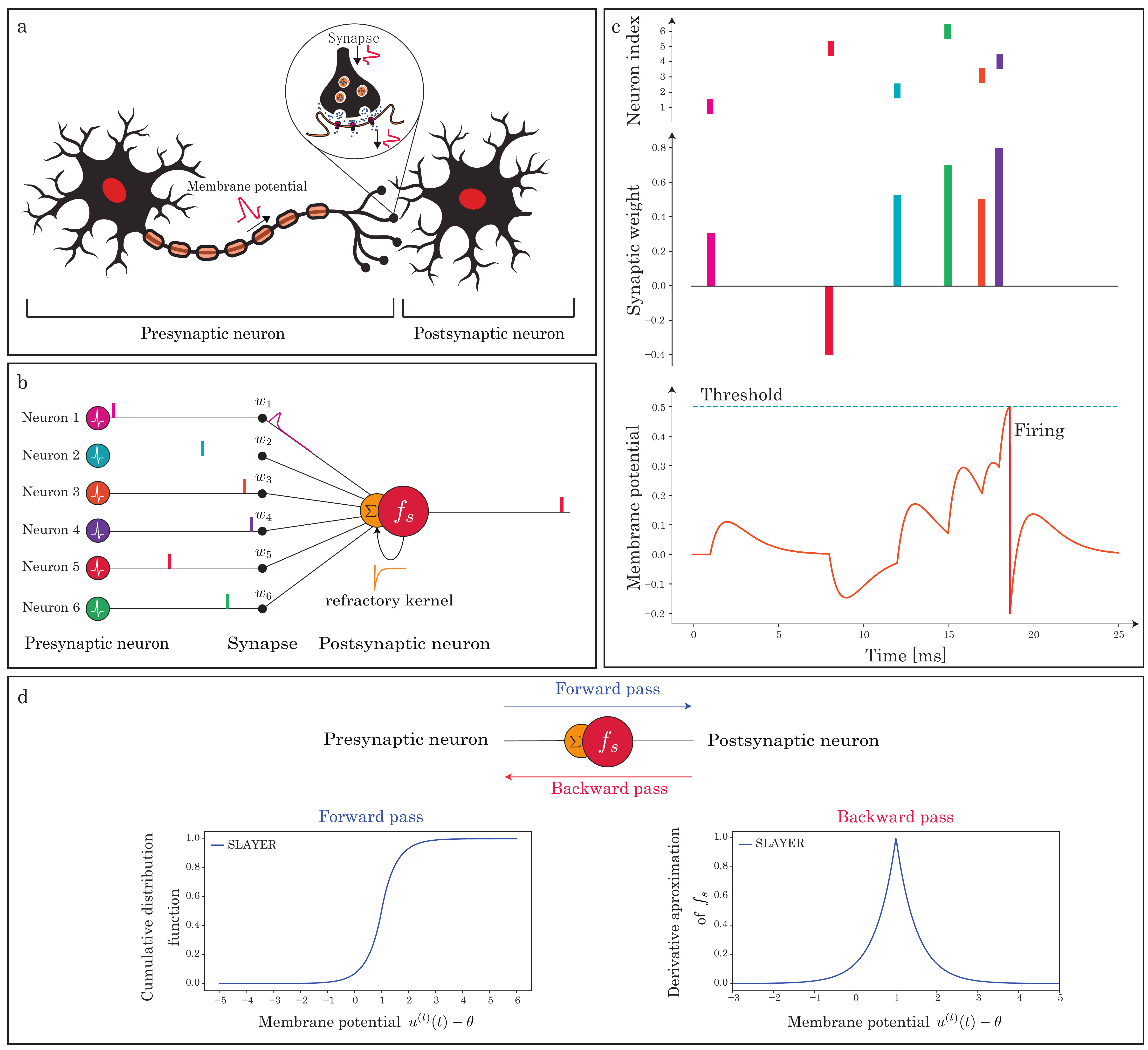}
\caption{Illustration of fundamental components of SNN. (a) Biological information and signal processing between presynaptic and postsynaptic neurons in SNN. (b) A spiking neuron processes and communicates binary spiking events over time. (c) The postsynaptic neuron changes the membrane potential through each layer of the post-synaptic potential (PSP). It generates the output spikes when the membrane potential reaches the neuronal firing threshold~\cite{Comsa-2020}. (d) The cumulative distribution function (``Forward pass'') of our surrogate gradient and itself (``Backward pass'')~\cite{Shrestha-2018}.}
\label{fig:label1}
\end{center}
\end{figure} 

%%%%%%%%%%%%%%%%%%%%%%%%%%%%%%%%%%%%%%%%%%
\section{Related Works}
\label{sec:related-works}

\subsection{Spiking Neuron}
SNN consists of spiking neurons that model the behavior of biological neurons and handle the firing timing of the spikes. Owing to the differences in approximations, several spiking neuron models have been proposed, such as the integrate-fire (IF) \cite{Lapique-1907}, leaky-integrate-and-fire (LIF) \cite{Stein-1965}, Izhikevich \cite{Izhikevich-2003}, and Hodgkin-Huxley model \cite{Hodgkin-1952}. In this study, we adopt the spike response model (SRM) \cite{Gerstner-2002} to deal with the refractory period (Section \ref{sec:2}). 

The refractory period is an essential function of biological neurons to suppress the spike firing. Spike firing occurs when the neuron's membrane potential exceeds the firing threshold. 
From a biological perspective, the membrane potential is calculated using PSP, representing the electrical signals converted from the chemical signals. These behaviors are represented within the chemical synapse model shown in Figure \ref{fig:label1} (a) \cite{Rall-1967}. SRM was implemented to approximate this synaptic model better than IF/LIF neurons, which are widely used in SNNs.

\subsection{Training of Spiking Neural Networks}
It is well-known that SNNs are difficult to train due to non-differential spike trains. Researchers are working on this problem, and their solutions can be divided into two approaches: first, the ANN-SNN conversion \cite{Diehl-2015b, Rueckauer-2017, Sengupta-2019, Li-2021}, and second, the usage of the surrogate gradient \cite{Wu-2018, Zenke-2018, Shrestha-2018, Lee-2020, Zhang-2020}. The ANN-SNN conversion method uses the trained ANN parameters of SNN. The sophisticated and state-of-the-art ANN model can be reused through this method. However, this conversion approach requires many time-steps during inference and increase the power consumption. In contrast, the surrogate gradient is used to directly train SNNs by approximating the gradient. The surrogate gradient approach was adopted since the model obtained by surrogate gradient requires far fewer inference time-steps than the ANN-SNN conversion model \cite{Diehl-2015a}. 

\subsection{Normalization}
One of the techniques that have contributed to the success of ANNs is Batch Normalization (BN) \cite{Ioffe-2015}. BN is used to reduce the internal covariate shift, leading to a smooth landscape \cite{Santurkar-2018} while corresponding to the homeostatic plasticity mechanism of BNNs \cite{Shen-2021}. Using a mini-batch, BN computes the sample mean and standard deviation (STD). Meanwhile, several variants have been proposed to compute the sample mean and STD, such as Layer Normalization (LN) \cite{Ba-2016}, Instance Normalization (IN) \cite{Ulyanov-2016}, and Group Normalization (GN) \cite{Wu-Yuxin-2018}. In particular, LN is effective at stabilizing the hidden state dynamics in recurrent neural networks for time-series processing \cite{Ba-2016}. 

\begin{figure}[t]
\begin{center}
\includegraphics[width=1.0\linewidth]{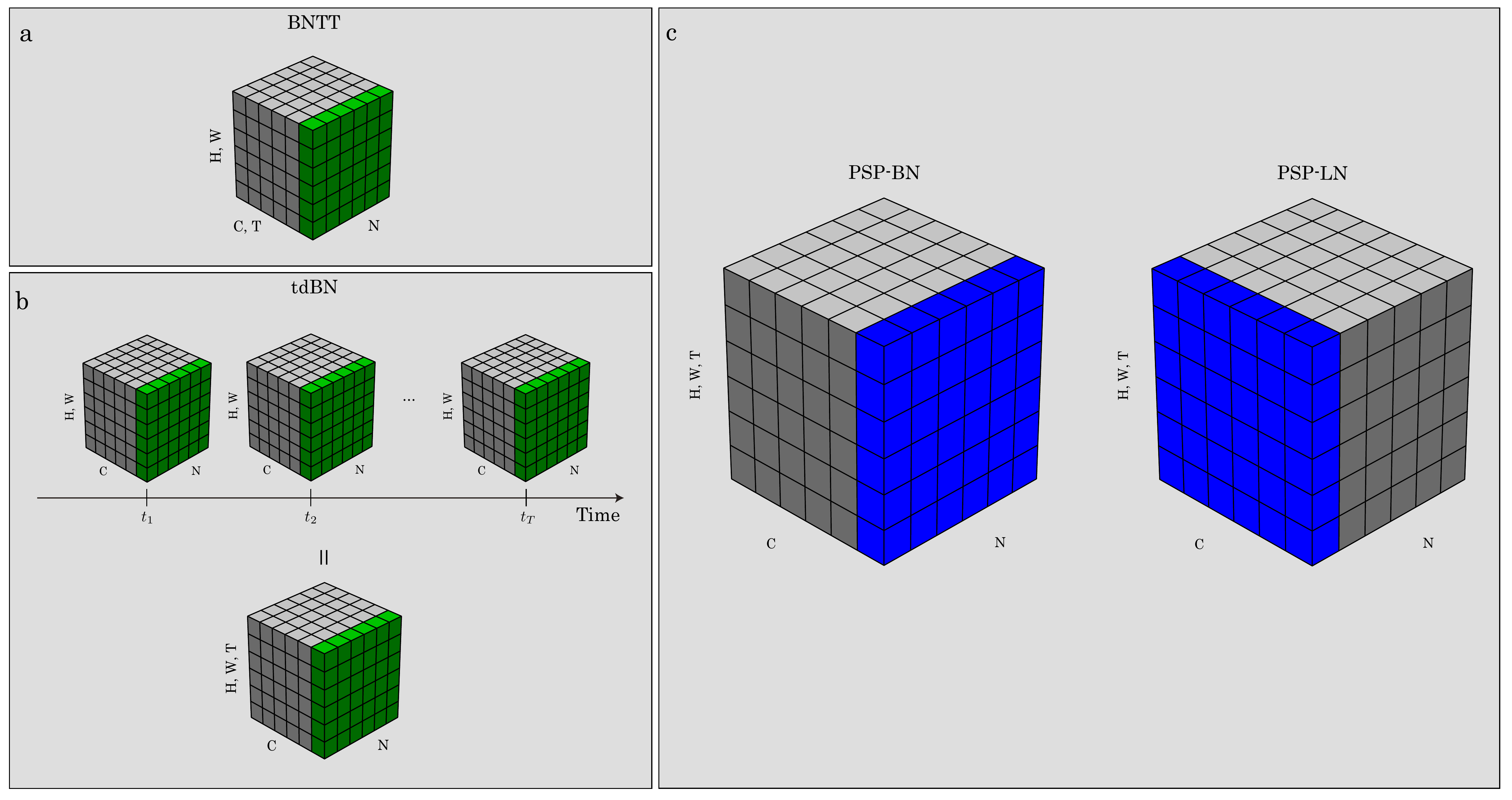}
\caption{Normalization methods for spiking neural networks (SNNs). Each subplot shows a feature map (a, b) or post-synaptic input (b) tensor, with N as the batch axis, C as the channel axis, (H, W) as the spatial axis, and T as the time axis in the figure. Green voxels (a, b), previous methods, and blue (b), our proposed method, are normalized by the same second central moment and uncentered second moment, respectively.}
\label{fig:label23}
\end{center}
\end{figure}

Several normalization methods have also been proposed in the field of SNNs, such as threshold-dependent BN (tdBN) \cite{Zheng-2020} and BN through time (BNTT) \cite{Kim-2020}. Thus, tdBN incorporates the firing threshold into BN, whereas BNTT computes BN at each time step. Furthermore, some studies used BN as is \cite{Ledinauskas-2020}. These studies applied the normalization to the membrane potential. In contrast, our method was applied to PSP, as shown in Figure \ref{fig:label23} (b), to simplify the normalization form (Section \ref{sec:3}).

%%%%%%%%%%%%%%%%%%%%%%%%%%%%%%%%%%%%%%%%%%
\section{Spiking Neural Networks based on Spike Response Model}
\label{sec:2}
In this section, we describe the SNN used in this study. Our SNN is constructed using SRM \cite{Gerstner-2002}; it uses SLAYER \cite{Shrestha-2018} as the surrogate gradient function to train the SRM.

\subsection{Spike Response Model}
We adopt SRM as a spiking neuron model \cite{Gerstner-2002}. SRM model is based on combining the effects of the incoming spike arriving at the spiking neuron. It also has a function to the spike firing when the membrane potential $u(t) (t=1, 2, \cdots, T)$ reaches the firing threshold. Figures \ref{fig:label1} (b) and (c) indicate the behavior of this model. The equations are given as follows:
\begin{eqnarray}
  u_{i}(t) &=& \sum_{j}w_{ij}(\varepsilon\ast s_{j})(t)+(\nu\ast s_{i})(t), \label{eq:1} \\
  s_{i}(t) &=& f_s (u_{i}(t)-\theta ), \label{eq:2} 
\end{eqnarray}
where $w_{i, j}$ is the synaptic weight from the presynaptic neuron $j$ to the postsynaptic neuron $i$. $s_{j}(t)$ is the spike train inputted from the presynaptic neuron $j$, $s_{i}(t)$ is the output spike train of the postsynaptic neuron $i$, $\ast$ is a temporal convolution operator, and $\theta$ is a threshold used to control the spike generation. $f_s$ is the Heaviside step function, which fires the spike when the membrane potential $u_i(t)$ exceeds the firing threshold $\theta$. In addition, $\varepsilon (\cdot )$ and $\nu (\cdot )$ are the spike response and refractory kernels formulated using the exponential function as follows: 
\begin{eqnarray}
  \varepsilon (t) &=&\frac{t}{\tau_{s}}e^{1-\frac{t}{\tau_{s}}}, \label{eq:3} \\ 
  \nu (t) &=&-2\theta e^{-\frac{t}{\tau_{r}}}, \label{eq:4}  
\end{eqnarray}
where $\tau_{s}$ and $\tau_{r}$ are the time constants of spike response and refractory kernels, respectively. Note that $\varepsilon\ast s_{j}(t)$ represents the PSP.

\subsection{Multiple Layers Spike Response Model}
By using Equations (\ref{eq:1}) and (\ref{eq:2}), the SNNs with multi-layers can be described as follows:
\begin{eqnarray}
\displaystyle \bm{a}^{(l)}(t)&=&(\varepsilon\ast \bm{s}^{(l)})(t), \label{eq:5} \\
\displaystyle \bm{u}^{(l+1)}(t)&=&\bm{W}^{(l)}\bm{a}^{(l)}(t)+(\nu\ast\bm{s}^{(l+1)})(t), \label{eq:6} \\
\displaystyle \bm{s}^{(l+1)}(t)&=&f_{s}(\bm{u}^{(l+1)}(t)), \label{eq:7} 
\end{eqnarray}
where $\bm{a}^{(l)}(t) \in \mathbb{R}_{\geq 0}^{\mathrm{C}\times\mathrm{W}\times \mathrm{H}}$ and $\bm{s}^{(l)}(t)\in \{0,1\}^{\mathrm{C}\times\mathrm{W}\times \mathrm{H}}$ are the PSP and input spike tensor of time step $t$; $\mathrm{C}$ is the number of channels; and $\mathrm{W}$ and $\mathrm{H}$ are the width and height of the input spike tensor, respectively. Since $\bm{a}^{(l)}(t)$ does not take a value less than zero, we consider an excitatory neuron. Furthermore, $\bm{W}^{(l)} \in \mathbb{R}^M$ is the weight matrix representing the synaptic strengths between the spiking neurons in $l$ and $l+1$ layers; $M$ is the number of neurons of $l+1$-th layer. 

\subsection{Deep SNNs by Pre-activation Blocks}
Deep neural network are essential to recognize complex input patterns. In particular, ResNet is widely used in ANNs \cite{He-2016a, He-2016b}, and its use in SNNs is expanding. 

\begin{figure}[t]
%\widefigure
\centering
\includegraphics[width=1.0\linewidth]{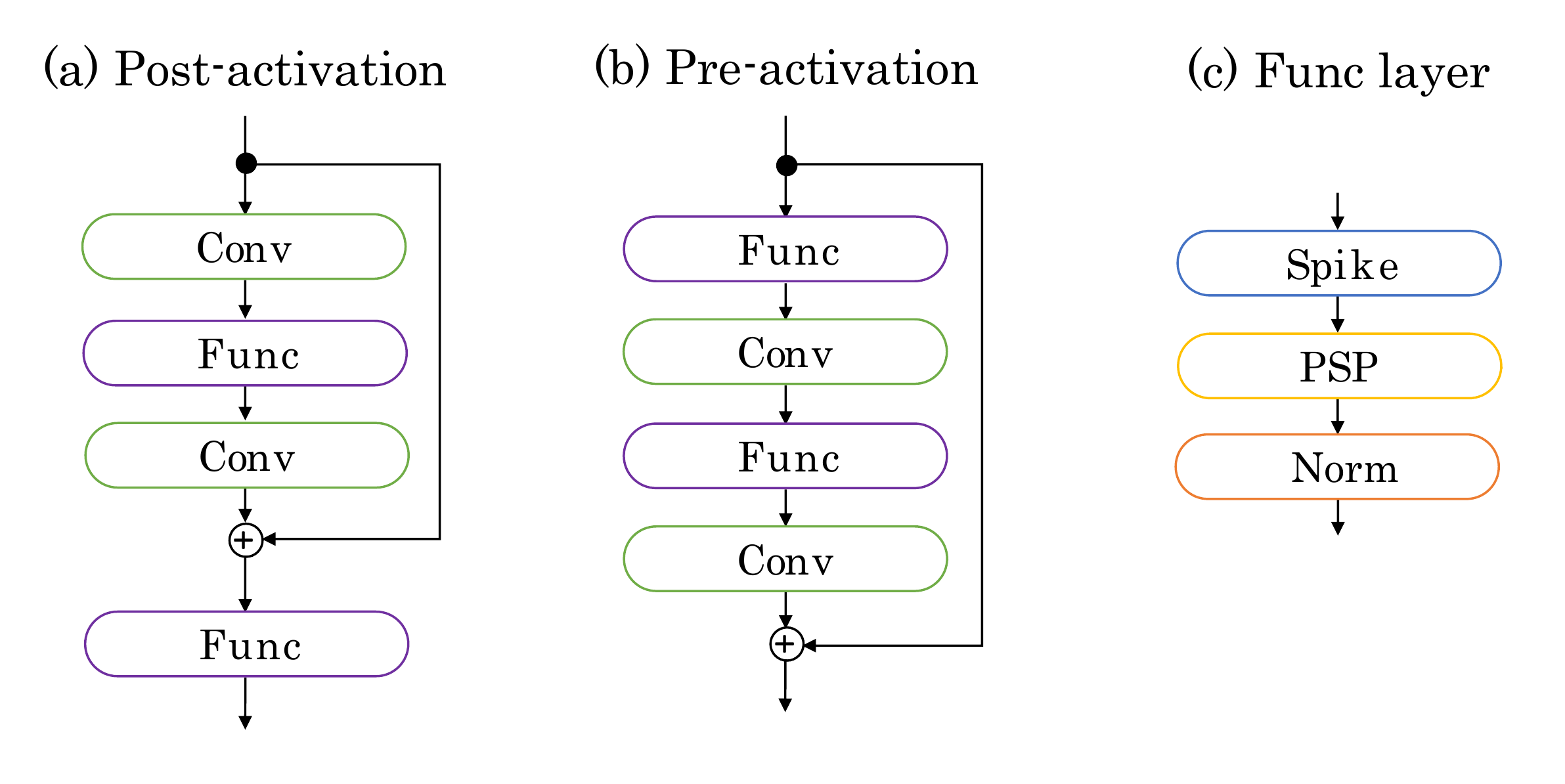}
%\includegraphics[width=11 cm]{3.pdf}
%\vspace{5zh}
\caption{Network blocks. (a) Post-activation residual block, (b) Pre-activation residual block, and (c) Func layer. Postsynaptic potential (PSP) and Norm in the Func layer represent Equation (\ref{eq:5}) and normalization. Spike represents $f_s(.+r)$, where $r$ is the refractory period.}
\label{fig:label33}
\end{figure}

The ResNet's networks are divided into the pre-activation and post-activation residual blocks as follows (Figure \ref{fig:label33}):
\begin{eqnarray}
\displaystyle {\sf Pre}: \bm{h}^{(k+1)}(t)&=&\bm{h}^{(k)}(t)+G(\bm{h}^{(k)}(t)), \label{eq:18} \\
\displaystyle {\sf Post}:
\bm{h}^{(k+1)}(t)&=& F(\bm{h}^{(k)}(t)+G(\bm{h}^{(k)}(t))), \label{eq:19}
\end{eqnarray}
where $\bm{h}^{(k)}$ and $\bm{h}^{(k+1)}$ are the input and output in the $k+1$ block, respectively. $G$ represents the residual function, corresponding to ``Conv-Func-Conv'' and ``Func-Conv-Func-Conv'' in Figure \ref{fig:label33}; $F$ represents the Func layer (``Spike-PSP-Norm''). Note that the refractory period is used in $F$. In the experimental section, we compare these blocks and show that deep SNNs can be trained using the pre-activated residual blocks. This result shows that identity mapping is an essential tool to train the deep SNNs, similar to ANNs \cite{He-2016b}.

\subsection{Surrogate-Gradient}
We use SLAYER \cite{Shrestha-2018} as one of the surrogate gradient algorithms to tarin the SNN with multi-layers. In SLAYER, the derivative of the spike activation function $f_{s}$ of the $l+1$ layer is approximated as follows (Figure \ref{fig:label1} (d)):
\begin{eqnarray}
\displaystyle \bm{\rho}^{(l+1)}(t)&=&\frac{1}{\alpha}(-\beta |\bm{u}^{(l+1)}(t)-\bm{\theta}|), \label{eq:8} 
\end{eqnarray}
where $\alpha$ and $\beta$ are hyperparameters, and $\bm{\theta} \in \mathbb{R}^M$ is the firing threshold. SLAYER can be used to train SRM as described in \cite{Shrestha-2018}.

\begin{figure*}[t]
%\widefigure
\begin{center}
\includegraphics[width=1.0\linewidth]{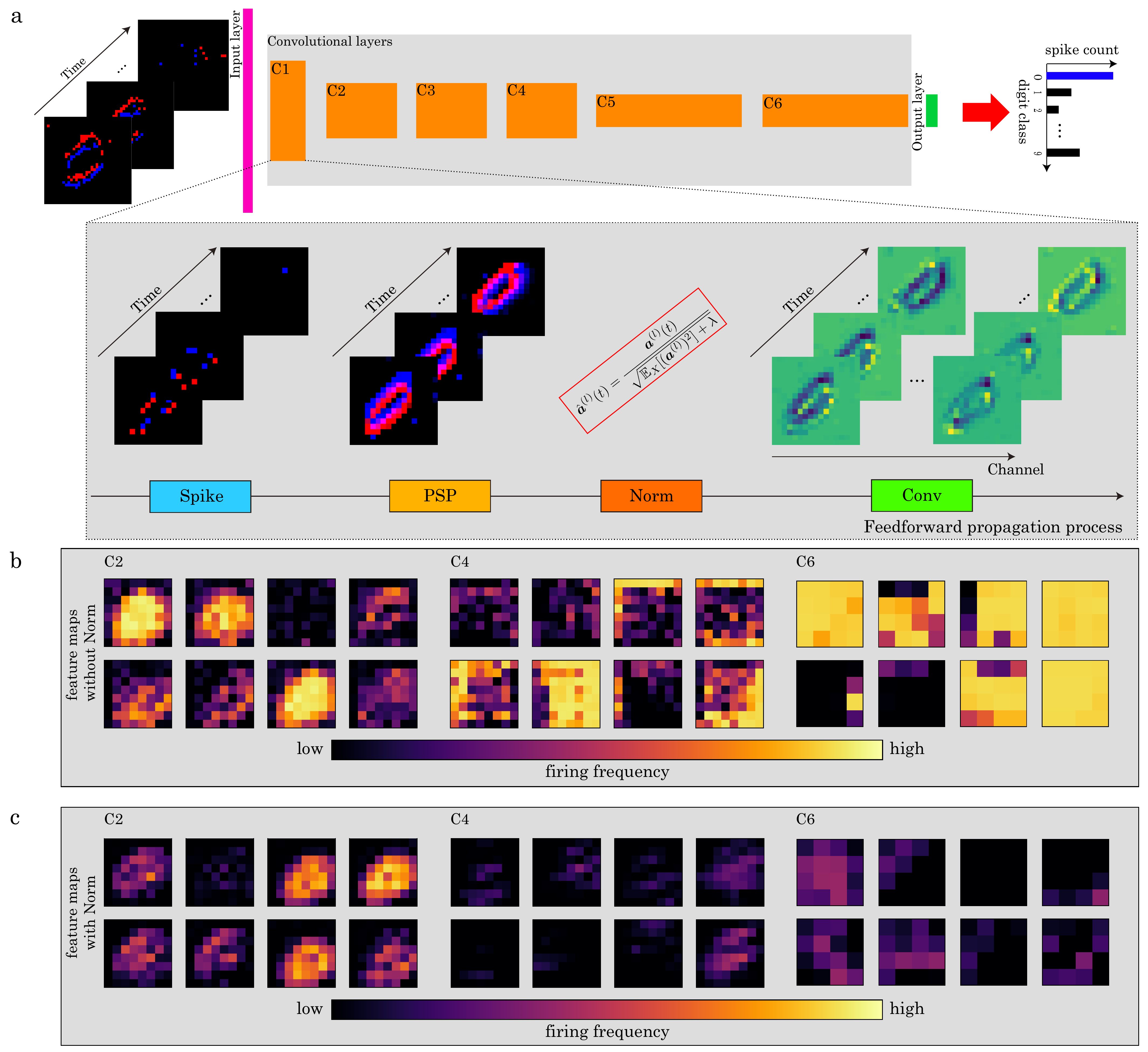}
%\includegraphics[width=13 cm]{4.pdf}
%\vspace{5zh}
\caption{(a) Overview of the forward propagation phase using postsynaptic potential (PSP) normalization on the N-MNIST dataset. (b) Example of feature maps, which are integrated along the time axis, without PSP normalization. (c) As in (b) but for feature maps with it. PSP normalization controls the activity of the network and prevents the over-firing of the neurons.}
\label{fig:label34}
\end{center}
\end{figure*} 

%%%%%%%%%%%%%%%%%%%%%%%%%%%%%%%%%%%%%%%%%%
\section{Normalization of Postsynaptic Potential}
\label{sec:3}

In this section, we explain the derivation of our normalization, which is called {\it postsynaptic-potential normalization,} as shown in Figure \ref{fig:label34} (a). 

As the depth of the SNN becomes deeper, it becomes more difficult to control spike firing properly (Figures \ref{fig:label34} (b) and (c)). To tackle this problem, we first introduce the following typical normalization into the PSP. 
\begin{eqnarray}
\displaystyle \bm{\hat{u}}^{(l+1)}(t)&=&\bm{W}^{(l)}\bm{\hat{a}}^{(l)}(t)+(\nu\ast\bm{s}^{(l+1)})(t), \label{eq:9} \\
\displaystyle \bm{\hat{a}}^{(l)}(t)&=&\frac{\bm{a}^{(l)}(t)-\mathbb{E}_X[\bm{a}^{(l)}]}{\sqrt{\mathbb{V}_X[\bm{a}^{(l)}]+\lambda}}\odot \gamma +\xi, \label{eq:10}
\end{eqnarray}
where $\gamma$ and $\xi$ are trainable parameters; each variable of $\mathbb{E}_X[\bm{a}^{(l)}]$ and $\mathbb{V}_X[\bm{a}^{(l)}]$ is approximate as follows:
\begin{eqnarray}
\displaystyle \mathbb{E}_X[a^{(l)}_i]&\approx&\frac{1}{X}\sum_{x=1}^{X} a^{(l)}_{i}(x), \label{eq:11} \\
\displaystyle \mathbb{V}_X[a^{(l)}_i]&\approx&\frac{1}{X}\sum_{x=1}^{X}(a^{(l)}_{i}(x)- \mathbb{E}_X[a^{(l)}_i])\odot (a^{(l)}_{i}(x,t) -\mathbb{E}_X[a^{(l)}_i]), \label{eq:12}
\end{eqnarray}
where $a^{(l)}_{i}(x)$ represents the $x$-th variable required to compute these statistics of the $i$-th variable of $\bm{a}^{(l)} \in \mathbb{R}_{\geq 0}^{\mathrm{C}\times\mathrm{W}\times \mathrm{H} \times \mathrm{N} \times \mathrm{T}}$ ($N$ is the mini-batch size), and $X$ depends on what kind of summation to compute. For example, if we compute these equations as in BN, $X = W \times H \times N \times T$. In addition, if we compute them as in LN, $X = W \times H \times C \times T$. Note that the normalization to PSP means that it inserts before the convolution or fully connected layers. This position differs from the other normalization ones, which use normalization to the membrane potential \cite{Ledinauskas-2020,Kim-2020,Zheng-2020}.

As shown in Equation (\ref{eq:10}), $\bm{\hat{a}}^{(l)}(t)$ may take minus. Therefore, $\bm{\hat{a}}^{(l)}(t) < 0$ is not valid since neurons of SLAYER represent excitatory neurons. This phenomenon clearly arises from the trainable parameter $\xi$ and the shift parameter $\mathbb{E}_X[\bm{a}^{(l)}]$. Thus, we modify Equation (\ref{eq:10}) as follows:

\begin{eqnarray}
\displaystyle \bm{\hat{a}}^{(l)}(t)&=&\frac{\bm{a}^{(l)}(t)}{\sqrt{\mathbb{V}_X[\bm{a}^{(l)}]+\lambda}}\odot\gamma. \label{eq:13} 
\end{eqnarray}
Next, we consider the case when $\bm{\hat{u}}^{(l+1)}(t)$ reaches the firing threshold $\bm{\theta}$.
\begin{eqnarray}
\displaystyle \bm{\theta}&=&\bm{W}^{(l)}\bm{\hat{a}(t)}+(\nu\ast\bm{s}^{(l+1)})(t), \label{eq:14} \\
\displaystyle &=&\bm{\hat{W}}^{(l)}\frac{\bm{a}^{(l)}(t)}{\sqrt{\mathbb{V}_X[\bm{a}^{(l)}]+\lambda}}+(\nu\ast\bm{s}^{(l+1)})(t). \label{eq:15} 
\end{eqnarray}
Here, we merged the trainable parameter $\gamma$ and the weight matrix $\bm{W}^{(l)}$ into $\bm{\hat{W}}^{(l)}$. This merging is possible because of the normalization performed before multiplying $\bm{W}^{(l)}$. Then, we express Equation (\ref{eq:15}) as follows:
\begin{eqnarray}
\displaystyle \bm{\hat{W}}^{(l)}\bm{a}^{(l)}(t)&=&\sqrt{\mathbb{V}_X[\bm{a}^{(l)}]+\lambda}(\bm{\theta}-\nu\ast\bm{s}^{(l+1)}(t)) \ \ \ \coloneqq \ \ \ \hat{\bm{\theta}}. \label{eq:16} 
\end{eqnarray}
Equation (\ref{eq:16}) shows that the firing threshold varies dynamically, which is consistent with the activity of cortical neurons in the human brain \cite{{Frankenhaeuser-1965}, {Schlue-1974}, {Stafstrom-1984}, {Aertsen-1996}}. The refractory period $(\nu\ast\bm{s}^{(l+1)})(t)$ and $\sqrt{\mathbb{V}_X[\bm{a}^{(l)}]+\lambda}$ can decrease $\hat{\bm{\theta}}$ and scaling, respectively. 

Next, we focus on the scale factor $\sqrt{\mathbb{V}_X[\bm{a}^{(l)}]+\lambda}$. As shown in Equation (\ref{eq:16}), the firing threshold $\hat{\bm{\theta}}$ becomes larger as the variance (second central moment) $\mathbb{V}_X[\bm{a}^{(l)}]$ increases. However, considering the behavior of the membrane potential, $\hat{\bm{\theta}}$ should become larger when the value of PSP (not variance) increases. Thus, we modify the equation as follows.

\begin{eqnarray}
\displaystyle \bm{\hat{a}}^{(l)}(t)&=&\frac{\bm{a}^{(l)}(t)}{\sqrt{\mathbb{E}_X[(\bm{a}^{(l)})^2]+\lambda}}, \label{eq:17}
\end{eqnarray}
where $\mathbb{E}_X[(\bm{a}^{(l)})^2]$ represents the second raw moment consisting of the following variable,
\begin{eqnarray}
\displaystyle \mathbb{E}_X[(a^{(l)}_i)^2]&\approx&\frac{1}{X}\sum_{x=1}^{X} a^{(l)}_{i}(x) \odot a^{(l)}_{i}(x). \label{eq:expectation}
\end{eqnarray}
By using this equation, we do not have to compute the mean beforehand, in contrast to using the variance. 

\begin{figure}[t]
%\widefigure
\includegraphics[width=1.0\linewidth]{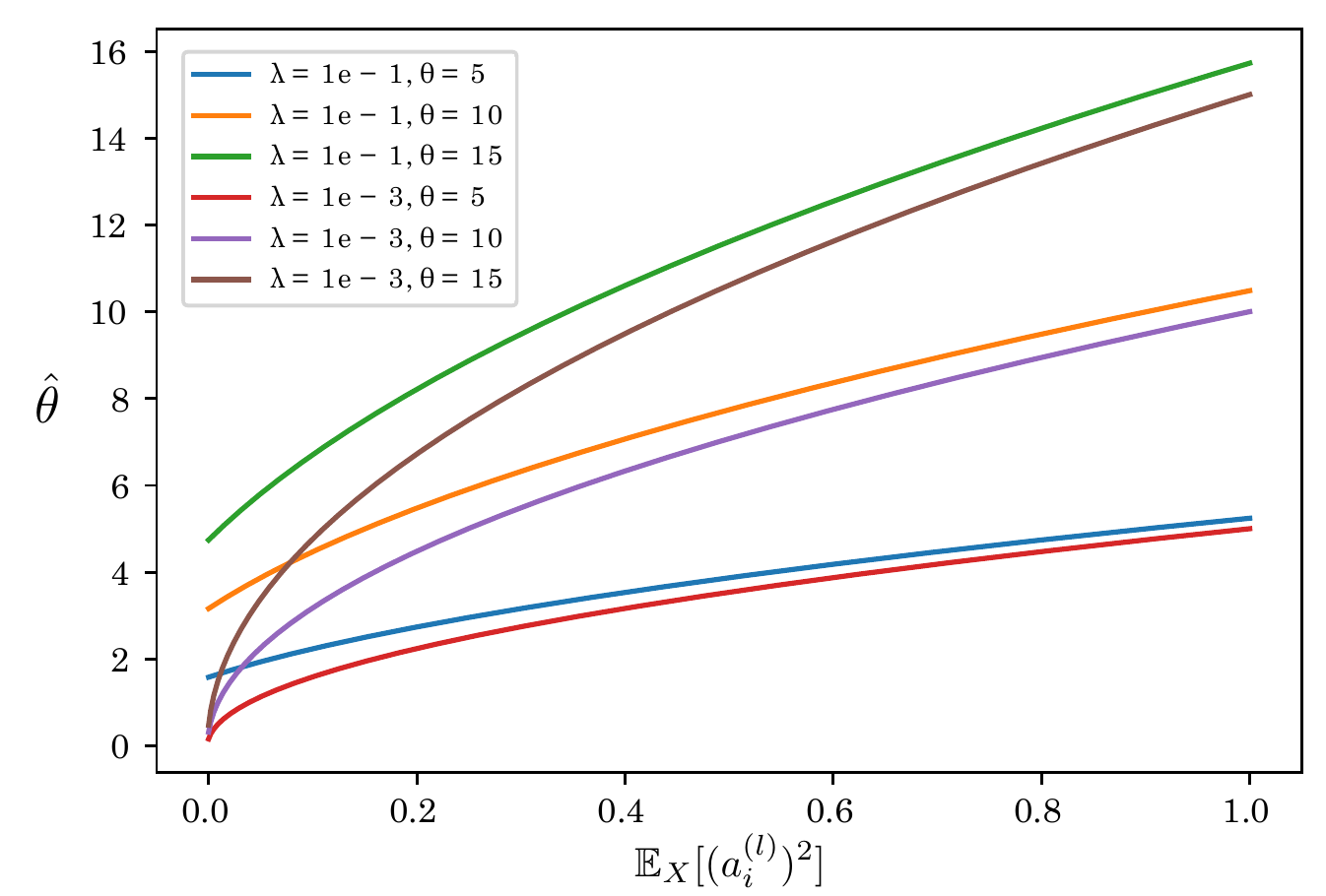}
\caption{Relationship between $\mathbb{E}_X[(a^{(l)}_i)^2]$ and $\hat{\theta}$.}
\label{fig:label5}
\end{figure}

In addition to $\mathbb{E}_X[(\bm{a}^{(l)})^2]$, there is a hyperparameter $\lambda$ in the scale factor. $\lambda$ is usually set to a small constant, e.g., $\lambda = 10^{-3}$ because it plays the role of the numerical stability. Figure \ref{fig:label5} shows the relationship between $\mathbb{E}_X[(a^{(l)}_i)^2]$ and $\hat{\theta}$ when changing $\theta$ and $\lambda$. As shown in this figure, $\hat{\theta}$ monotonically decreases as $\mathbb{E}_X[(a^{(l)}_i)^2]$ decreases. In particular, $\hat{\theta}$ is close to zero when $\lambda$ is sufficiently small, regardless of the initial threshold $\theta$. $\hat{\theta} \approx 0$ means that spikes fire at all times even if the membrane potential is significantly small, making it difficult to train a proper model. Thus, we set a relatively large value ($\lambda = 0.1$) as the default value.

\section{Experiments}
\label{sec:4}
In this section, we evaluate two PSP normalizations: BN (the most common normalization) and LN (which is effective in time-series processing, such as SNN). We called them {\it PSP-BN} ($X = W \times H \times N \times T$) and {\it PSP-LN} ($X = W \times H \times C \times T$). 

\subsection{Experimental Setup}
We evaluated PSP-BN and PSP-LN on the spatio-temporal event and static image datasets. We used N-MNIST \cite{Orchard-2015} and F-MNIST \cite{Xiao-2017}. N/F-MNISTs are widely used datasets containing 60K training and 10K test samples with 10 classes. Each size is $34 \times 34 \times 30000$ events (N-MNIST), and $28 \times 28$ pixels (F-MNIST). We partitioned the 60K data using 54K and 6K as our training and validation data, respectively. We also resized the F-MNIST image from $28 \times 28$ to $34 \times 34$ to achieve higher accuracy.

We evaluated the performance of several spiking convolutional neural network models, such as six-convolutional layers on both datasets and 14-convolutional layers on N/F-MNIST. We also used more deep models, such as ResNet-106 on N-MNIST and F-MNIST, respectively. 

We used hyperparameters shown in Table \ref{table1} in all experiments and implemented by PyTorch. We used the default initialization of PyTorch and showed the best accuracies of all models. All experiments were conducted using a single Tesla V100 GPU. In addition to this computational resource limitation, we randomly sampled 6K of the training data for both datasets to train in each epoch since SLAYER requires a significant amount of time to train. 

\begin{table}[t]
\begin{center}
\begin{tabular}{lcc}\bhline{1.2pt}
 Hyperparameter & N-MNIST & F-MNIST  \\ \hline
 $\tau_{s}$ & 10 & 10 \\
 $\tau_{r}$ & 10 & 10 \\
 $\alpha$ & 10 & 10 \\
 $\beta$ & 10 & 10 \\
 $\theta$ & 10 & 10 \\
 optimizer & AdaBelief & AdaBelief \\
 learning rate & $10^{-2}$ & $10^{-2}$ \\
 weight decay & $10^{-4}$ & $10^{-4}$ \\
 weight scale & 10 & 10 \\
 mini-batch size & 10 & 10 \\
 time step & 300 & 100  \\
 epoch & 100 & 100  \\ \bhline{1.2pt}
\end{tabular}
\caption{Hyperparameter setting on N-MNIST and F-MNIST.}
\label{table1}
\end{center}
\end{table}

\subsection{Effectiveness of Postsynaptic Potential Normalization}
We first evaluate the effectiveness of our normalizations. Table \ref{tab:2} presents the accuracies of PSP-BN and PSP-LN and other approaches. Note that we set our normalization before the convolution as described in Section \ref{sec:3}, which is different from the position proposed in previous studies \cite{Ledinauskas-2020,Kim-2020,Zheng-2020}. This table illustrates that PSP-BN and PSP-LN achieve high accuracies on both datasets compared to the other approaches. 

\begin{figure}[t]
%\widefigure
\includegraphics[width=1.0\linewidth]{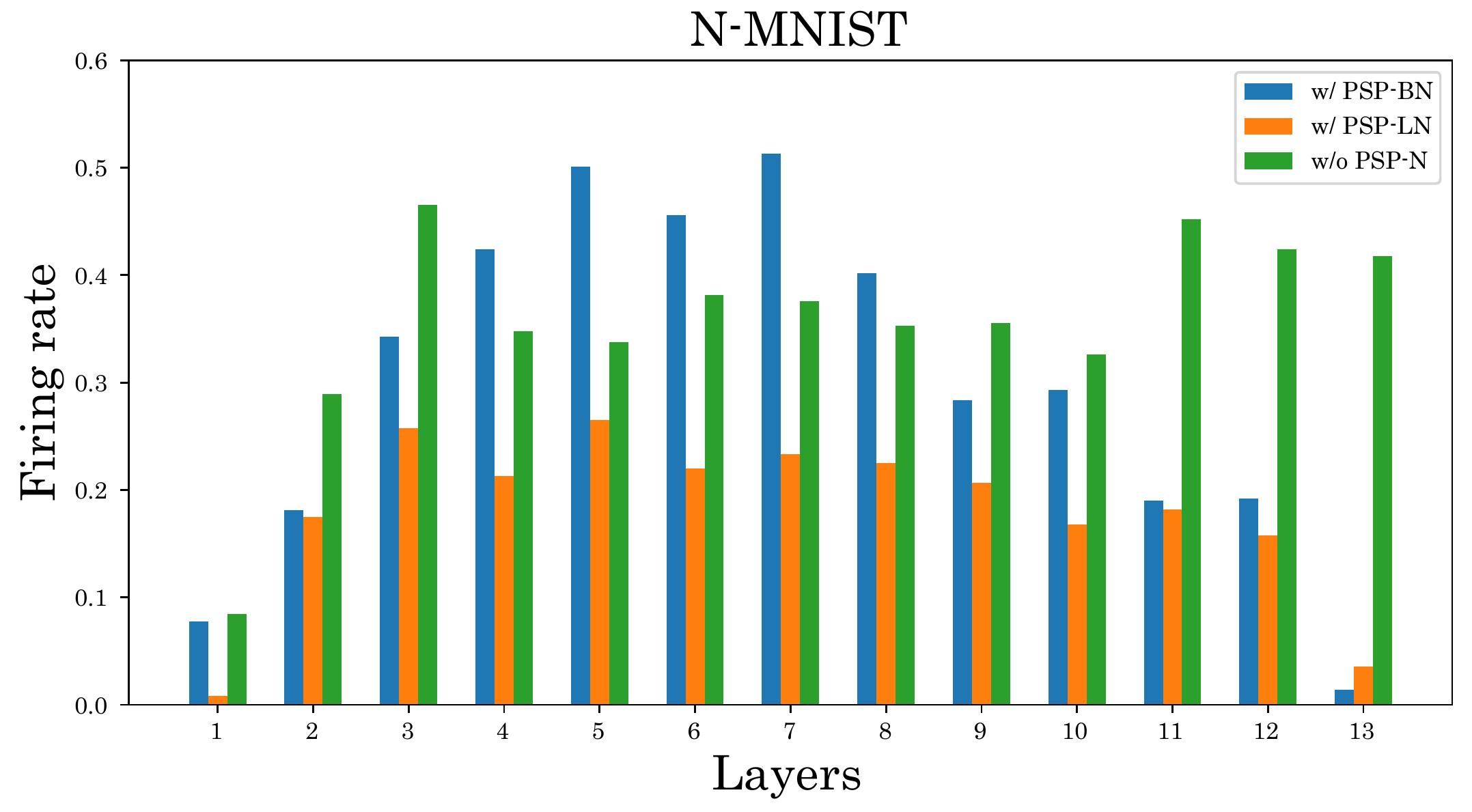}\\
\includegraphics[width=1.0\linewidth]{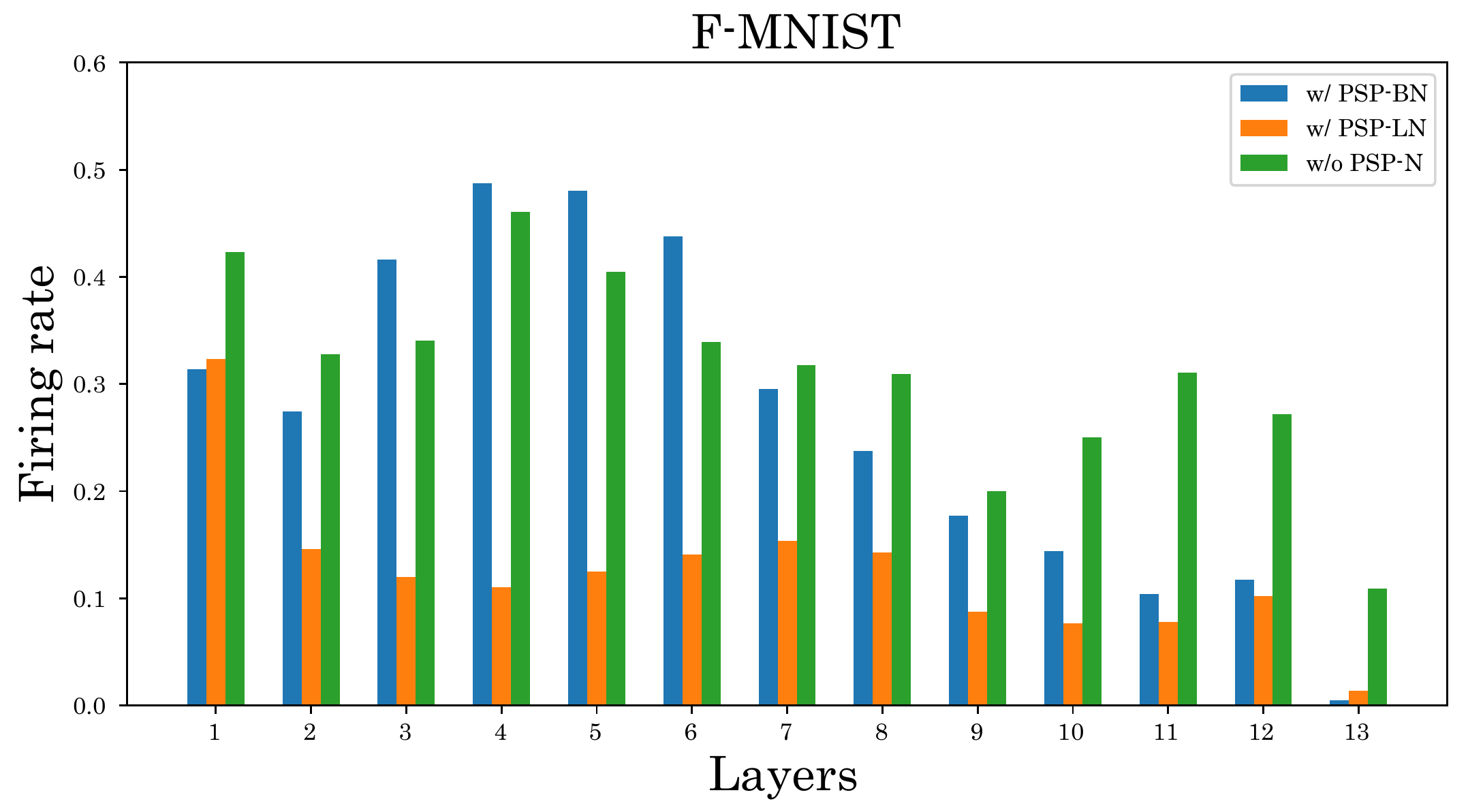}
\caption{Comparison of firing rates. The top is N-MNIST and the bottom is F-MNIST.}
\label{fig:firing_rate_layers}
\end{figure}

\begin{figure}[t]
%\widefigure
\includegraphics[width=1.0\linewidth]{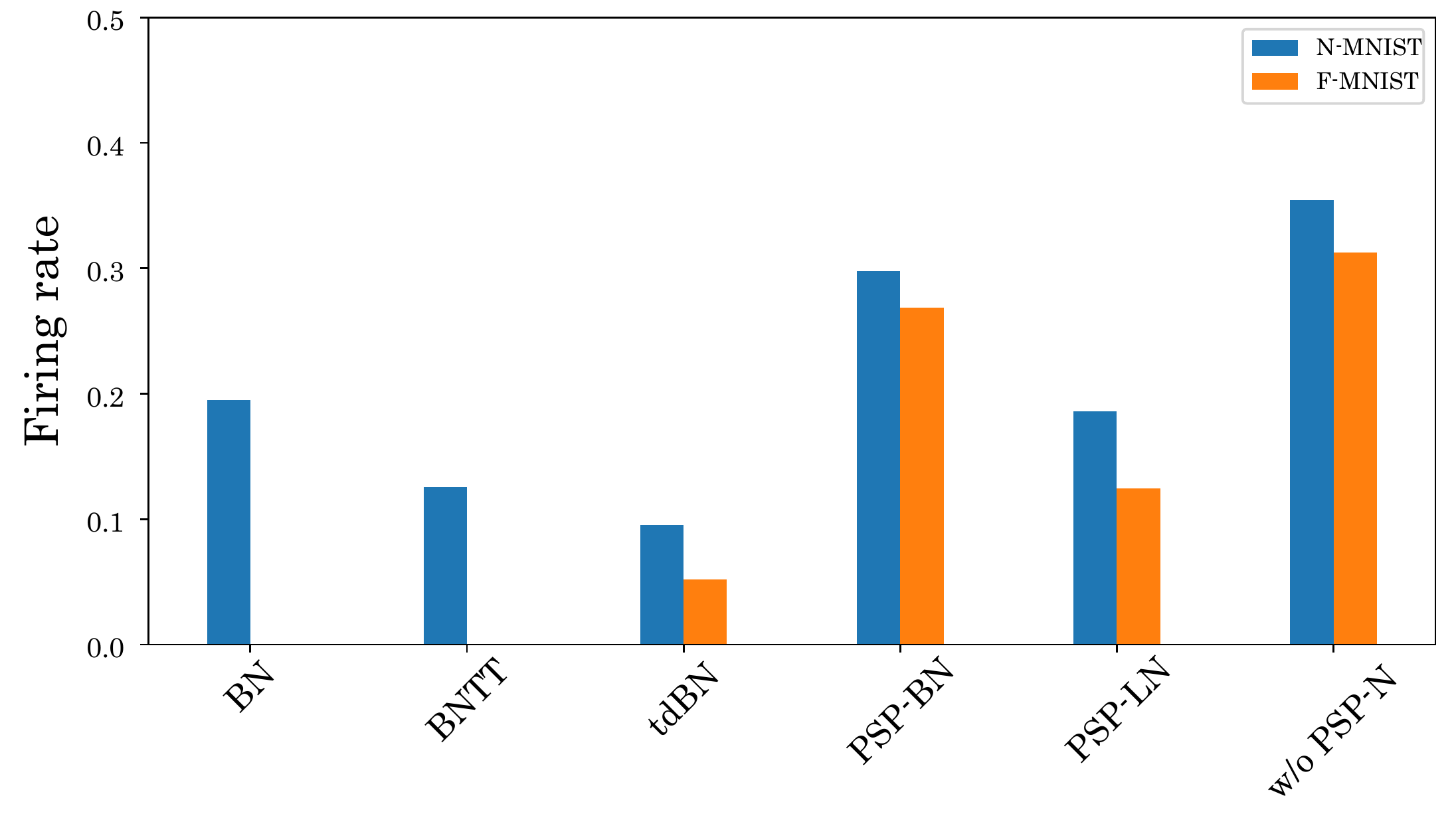}
\caption{Total firing rates of each model.}
\label{fig:firing_rate_all}
\end{figure}

We also investigate the effect of the proposed method on the firing rate. Figures \ref{fig:firing_rate_layers} and \ref{fig:firing_rate_all} show the firing rates of each method. As shown in Figure \ref{fig:firing_rate_layers}, our normalized models can suppress the firing rate in most layers compared to the unnormalized model. Furthermore, from Figure \ref{fig:firing_rate_all} and Table \ref{tab:2} show that our normalized models can simultaneously achieve high classification accuracy and low firing rate compared to other normalizations. These results verifies the effectiveness of our normalizations.

\begin{table}[t]
\begin{center}
\begin{tabular}{llll}
\bhline{1.2pt}
Method & Dataset & Network architecture & Acc. (\%)   \\ \hline
 BN \cite{Ledinauskas-2020} & N-MNIST & 34$\times$34$\times$2-8c3n-\{16c3n\}*5-16c3n-\{32c3n\}*5-10 & 85.1 \\
 BNTT \cite{Kim-2020} & N-MNIST  & 34$\times$34$\times$2-8c3n-\{16c3n\}*5-16c3n-\{32c3n\}*5-10 & 90.0 \\ 
 tdBN \cite{Zheng-2020} & N-MNIST  & 34$\times$34$\times$2-8c3n-\{16c3n\}*5-16c3n-\{32c3n\}*5-10  & 81.8 \\ 
 \textbf{PSP-BN} & N-MNIST  & 34$\times$34$\times$2-n8c3-\{n16c3\}*5-n16c3-\{n32c3\}*5-10 & 97.4 \\ 
 \textbf{PSP-LN} & N-MNIST  & 34$\times$34$\times$2-n8c3-\{n16c3\}*5-n16c3-\{n32c3\}*5-10 & 98.2 \\
 None & N-MNIST & 34$\times$34$\times$2-8c3-\{16c3\}*5-16c3-\{32c3\}*5-10 & 40.6 \\ \hline 
 BN \cite{Ledinauskas-2020} & F-MNIST & 34$\times$34-16c3n-\{32c3n\}*5-32c3n-\{64c3n\}*5-10 & 10 \\
 BNTT \cite{Kim-2020} & F-MNIST  & 34$\times$34-16c3n-\{32c3n\}*5-32c3n-\{64c3n\}*5-10 & 10 \\ 
 tdBN \cite{Zheng-2020} & F-MNIST  & 34$\times$34-16c3n-\{32c3n\}*5-32c3n-\{64c3n\}*5-10 & 40.5 \\ 
 \textbf{PSP-BN} & F-MNIST  & 34$\times$34-n16c3-\{n32c3\}*5-n32c3-\{n64c3\}*5-10 & 88.6 \\ 
 \textbf{PSP-LN} & F-MNIST  & 34$\times$34-n16c3-\{n32c3\}*5-n32c3-\{n64c3\}*5-10 & 89.1 \\ 
 None & F-MNIST & 34$\times$34-16c3-\{32c3\}*5-32c3-\{64c3\}*5-10 & 84.1 \\  \bhline{1.2pt}
\end{tabular}
\caption{Accuracies of N-MNIST and F-MNIST obtained from different methods. PSP-BN and PSP-LN are our normalization methods, and None is the model without normalization. Here, ``C'', ``n'', and ``d'' represent the convolution, normalization, and dense layers, respectively. In addition, each layer and spatial dimension in the network are separated by ``-'' and ``x''.}
\label{tab:2}
\end{center}
\end{table}

\subsection{Accuracy Dependency on Varying Hyperparameters}
\begin{figure}[t]
%\widefigure
\includegraphics[width=1.0\linewidth]{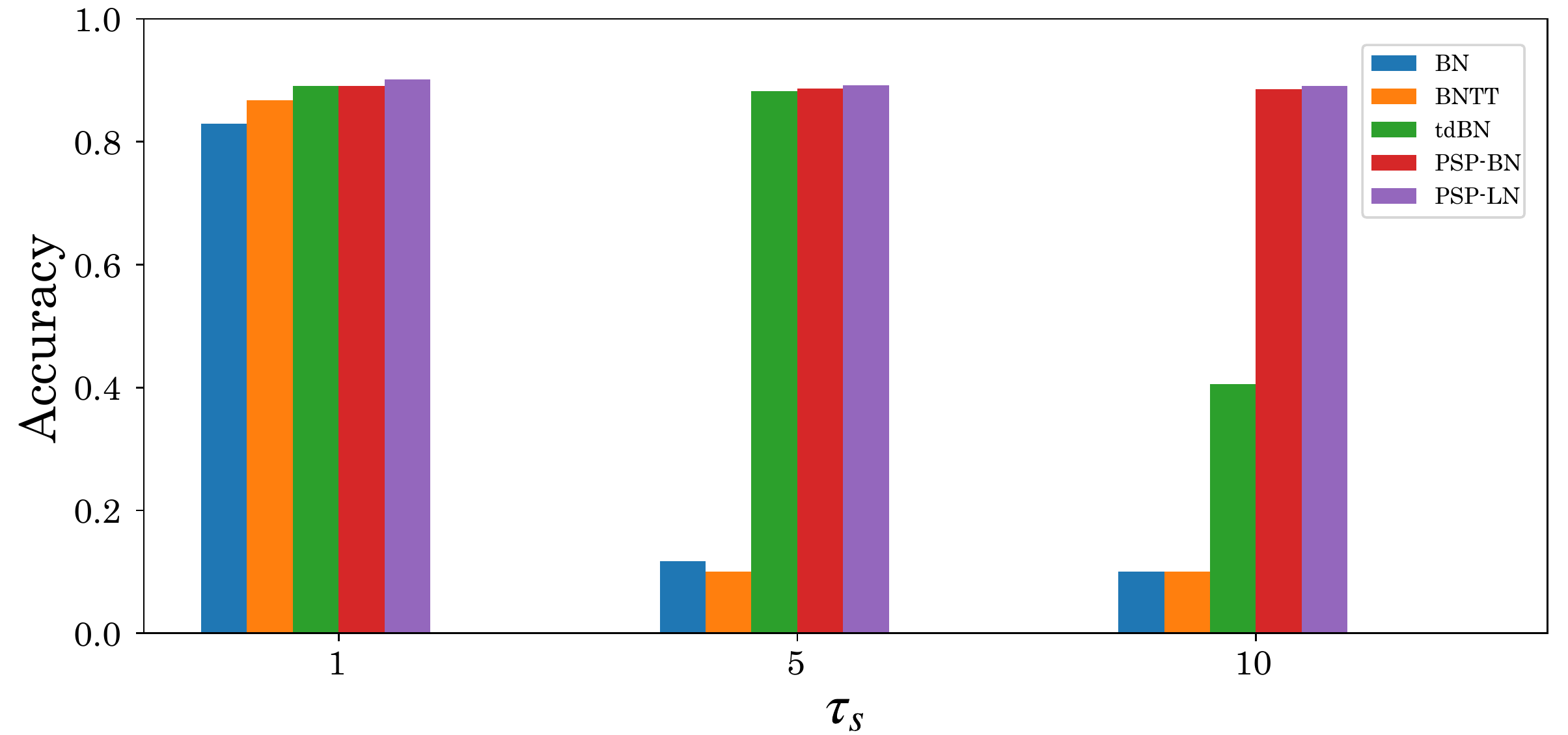}
\caption{Accuracy comparison of F-MNIST dataset with respect to changing hyperparameter $\tau_s$.}
\label{fig:accuracy_changing_tau}
\end{figure} 

Some authors pointed out that the tuning of hyperparameter $\tau_s$ is essential to achieve higher accuracy \cite{Fang-2021a}. Therefore, to investigate the effect of $\tau_s$, we compare the accuracy with respect to changing $\tau_s$. Figure \ref{fig:accuracy_changing_tau} shows F-MNIST's accuracies of each method for $\tau_s=\{1,5,10\}$. As shown in this figure, the models with PSP-BN/LN maintain the accuracy whereas others deteriorate the accuracies. This result shows that the model becomes robust to $\tau_s$ using our normalizations.

\begin{figure}[t]
%\widefigure
\includegraphics[width=1.0\linewidth]{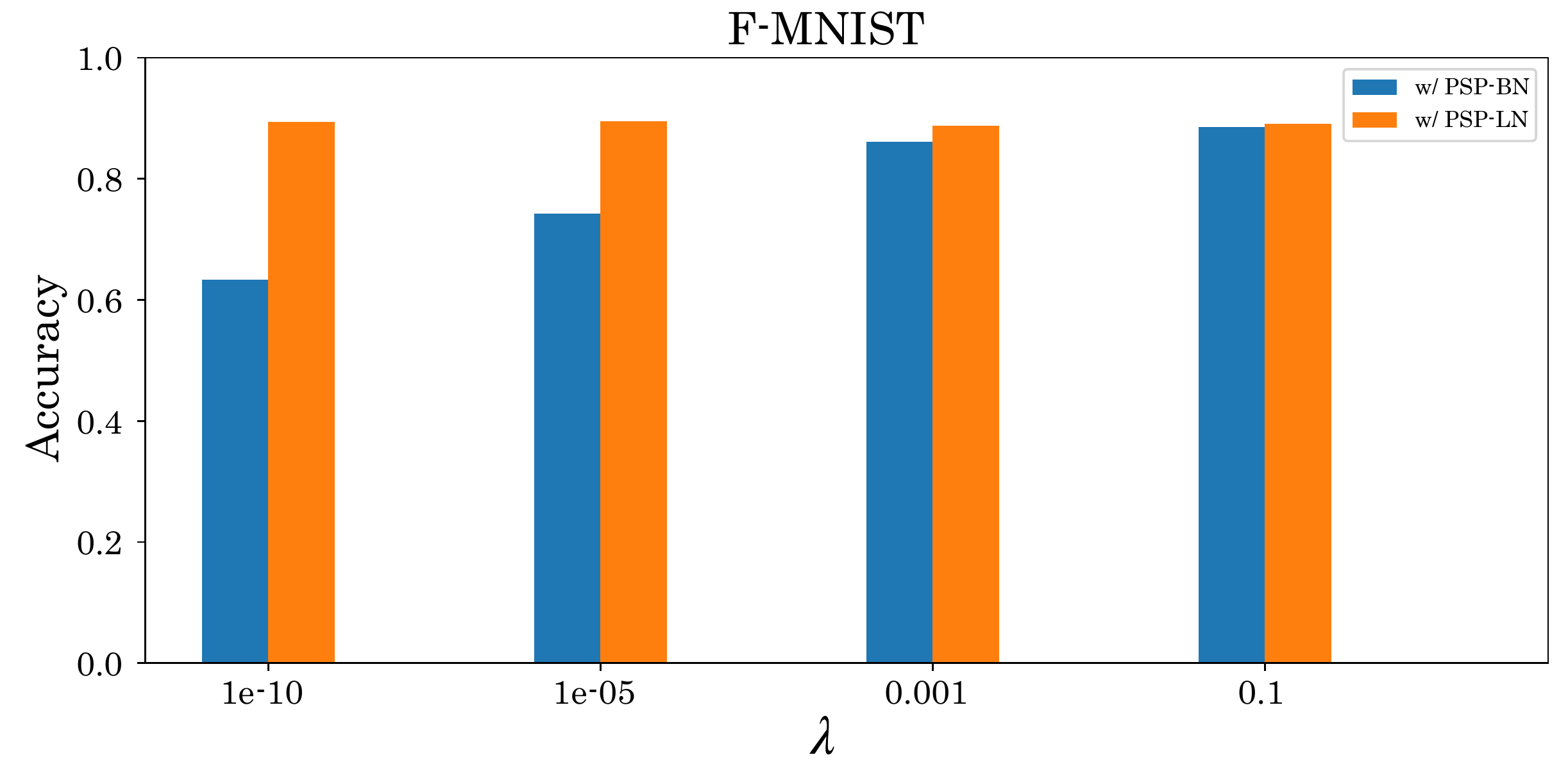}
\caption{Accuracy comparison of F-MNIST with respect to changing $\lambda$.}
\label{fig:accuracy_changing_lambda}
\end{figure}

In addition to $\tau_s$, we show the influence of $\lambda$. Figure \ref{fig:accuracy_changing_lambda} shows the performance when changing $\lambda$. As shown in this figure, the classification performance of PSP-BN becomes worse for a sufficient small value of $\lambda$. This result verifies the plausibility of our discussion in Section \ref{sec:3}.

\subsection{Performance Evaluation of Deep SNNs by Residual Modules}
\label{sub:resnet_result}

%\begin{table}[t]
\begin{table}[t]
\begin{center}
\begin{tabular}{lllc}\bhline{1.2pt}
 Meshod &Dataset & Network architecture & Acc. (\%) \\ \hline
 \textbf{PSP-BN} & N-MNIST & Post-activation ResNet-106 & 10.0 \\
 \textbf{PSP-BN} & N-MNIST & Pre-activation ResNet-106  & 75.4 \\ 
 \textbf{PSP-LN} & N-MNIST & Post-activation ResNet-106 & 10.0 \\
 \textbf{PSP-LN} & N-MNIST & Pre-activation ResNet-106  & 86.8 \\ \hline
 \textbf{PSP-BN} & F-MNIST & Post-activation ResNet-106 & 10.0 \\
 \textbf{PSP-BN} & F-MNIST & Pre-activation ResNet-106  & 81.6 \\ 
 \textbf{PSP-LN} & F-MNIST & Post-activation ResNet-106 & 10.0 \\
 \textbf{PSP-LN} & F-MNIST & Pre-activation ResNet-106  & 82.1 \\ \bhline{1.2pt} 
\end{tabular}
\caption{Performance comparison using post-activation and pre-activation residual blocks. We use ResNet-106 on N-MNIST and F-MNIST datasets, respectively.}
\label{tab:8}
\end{center}
\end{table}

Finally, we evaluate the performance of SNNs using the residual blocks. Table \ref{tab:8} shows the performance of SNNs using the pre-activation and post-activation residual blocks. As shown in this table, the accuracy is substantially improved using the pre-activation residual blocks. This result shows that the post-activation employed in previous studies without refractory period~\cite{Zheng-2020,Lee-2020,Fang-2021b} is unsuitable for SNNs with a refractory period. Thus, while ensuring the biological plausibility, due to the refractory period, we can obtain deep SNNs beyond 100 layers using our normalizations and pre-activation residual blocks.

\section{Conclusion}
\label{sec:5}
In this study, we proposed an appropriate normalization method for SNN. The proposed normalization removes the subtraction term from the standard normalization and uses the second raw moment as the denominator. Our normalized models outperformed other normalized models by inserting this simple normalization before the convolutional layer. Moewover, our proposed model with pre-activation residual blocks can train with more than 100 layers without any other special techniques dedicated to SNNs. In future studies, we will analyze more details and verify the effect on other datasets. Furthermore, we aim to extend postsynaptic normalization based on BN to develop robust normalization techniques for the thresholds in spiking neurons.

\end{document}